\pdfoutput=1

\documentclass[11pt]{article}

\usepackage{main}

\usepackage{times}
\usepackage{latexsym}

\usepackage[T1]{fontenc}

\usepackage[utf8]{inputenc}
\usepackage{enumitem}
\usepackage{microtype}
\usepackage{graphicx}
\usepackage{url}
\usepackage{amsmath}
\usepackage{multirow}
\usepackage{amsfonts}
\usepackage{booktabs}

\title{Update Frequently, Update Fast: Retraining Semantic Parsing Systems \\ in a Fraction of Time}
\author{
    Vladislav Lialin,\textsuperscript{\rm 1}
    \thanks{Work done while an intern at Google}
    Rahul Goel,\textsuperscript{\rm 2}
    Andrey Simanovsky,\textsuperscript{\rm 2}
    Anna Rumshisky,\textsuperscript{\rm 1}
    Rushin Shah\textsuperscript{\rm 2}
    \\
\textsuperscript{\rm 1}University of Massachusetts Lowell, 
\textsuperscript{\rm 2}Google\\ 
\texttt{vlialin@cs.uml.edu,
goelrahul@google.com,
asimanovsky@google.com,}
\\
\texttt{
arum@cs.uml.edu,
rushinshah@google.com}
}
\begin{document}

\maketitle

\begin{abstract}

Currently used semantic parsing systems deployed in voice assistants
can require weeks to train.
Datasets for these models often receive small and frequent updates, data patches. Each patch requires training a new model. To reduce training time, one can fine-tune the previously trained model on each patch, but na\"ive fine-tuning exhibits catastrophic forgetting -- degradation of the model performance on the data not represented in the data patch.

In this work, we propose a simple method that alleviates catastrophic forgetting and show that it is possible to match the performance of a model trained from scratch in less than 10\% of a time via fine-tuning. The key to achieving this is supersampling and EWC regularization. We demonstrate the effectiveness of our method on multiple splits of the Facebook TOP and SNIPS datasets.


\end{abstract}

\section{Introduction}

Semantic parsing is the task of mapping a natural language query into a formal language. Semantic parsing models are at the core of commercial dialogue systems like Google Assistant, Alexa, and Siri. Typically, for a given query, the model should identify the requested action (intent) and the associated values specifying parameters of the action (slots).
For example, if the query is \textit{Call Mary}, the action could be \textit{callAction} and \textit{Mary} could be a value of the slot \textit{contact}. 

Real-world datasets for semantic parsing get frequent updates, especially for the long tail of low-resource classes, and after each update, a new model needs to be trained.
Considering that such datasets can contain millions of examples \cite{damonte2019practical}, currently used neural semantic parsers can take days or even weeks of training.
To avoid training a model from scratch every time, we can fine-tune the current model on the new examples.
However, this leads to catastrophic forgetting \cite{french1999catastrophic} and to a performance drop on the original dataset.

Methods that mitigate catastrophic forgetting have been developed \cite{Kirkpatrick2017,Sun2020LAMOLLM}. However, there has been limited work studying their applicability to complex tasks such as semantic parsing \cite{lee2017toward} and to modern neural network models \cite{de2019episodic}. We also argue that a more practical data-incremental (as opposed to task-incremental) scenario has not received enough attention.

In this paper, we investigate the applicability of such methods to semantic parsing and demonstrate that using a simple combination of data supersampling and EWC \cite{Kirkpatrick2017} allows us to (1)~achieve approximately the same test-set performance as training a new model (2)~reduce catastrophic forgetting effect to almost negligible and (3)~to do so in less than 10\% of the time needed to train a model from scratch.
We also propose a new semantic parsing metric that is well-suited for per-class evaluation.

We evaluate our method in multiple settings, varying the amount of data available for each category initially and in subsequent increments.
We show that our method achieves higher accuracy than training from scratch in some of the more realistic settings, e.g., when less than 30 examples are available for a class in the original training data and less than 300 examples are added incrementally.

\section{Incremental training}
\label{sec:task}
We formalize our task as follows.
Suppose we have a network $\mathbb{M}_{prev}$ trained on the dataset $D_{1}$ as our current model.
We are then given a dataset $D_{2}$ 
as additional ``data patch'' data.
Let $T_1$ be the time to train
$\mathbb{M}_{from\_scratch}$ from scratch on $D_{1} \cup D_{2}$ and $T_2$ is the time
to fine-tune the previous model.
Our goal is to fine-tune $\mathbb{M}_{prev}$ and produce $\mathbb{M}_{from\_scratch}$,
resulting in comparable performance with $\mathbb{M}_{finetuned}$ with $T_2 \ll T_1$.
In this work, we  study the case where $D_2$ has a data distribution that is skewed and significantly different from the distribution in $D_1$.
Note that we do not explicitly study scenarios where conflicting training data is being corrected using $D_2$ as in \cite{gaddy2020overcoming}.

\textbf{Peculiarities of semantic parsing:} 
While standard classification tasks assume a single label for each training example, the output for semantic parsing is a tree with multiple nodes (Fig. \ref{fig:tree_paths}).
Tree structure can automatically account for the class correlations as the correlated classes are likely to occur in a single training example as different tree nodes.
Another potential benefit of having structured outputs is that the difference between the class distribution in $D_1$ and $D_2$ likely to be smaller than in unconstrained classification, since $D_2$ is likely to include some frequent classes from $D_1$ in its parsing trees.

Another characteristic of commercial semantic parsing is high number of classes. 
For example, the publicly available TOP dataset~\cite{gupta2018top} uses
25 intents and 36 slots for just two domains \textit{navigation} and \textit{events}.
Real production systems can contain hundreds of classes for a single domain \cite{rongali2020don}.
Thus, incremental training is particularly interesting as the long tail of the class distribution calls for frequent model updates.

\section{Fine-tuning approaches}
\label{sec:methods}

In continual learning, it is generally assumed that the previous data is not accessible at the next learning iteration \cite{de2019episodic}.
However, this is usually not the case for commercial semantic parsing systems.
Hence, we assume that we have access to all data.

Our main reason to apply continual learning methods to semantic parsing is to reduce training times. Thus, we only use the methods that do not require significant additional computation.
We investigate several fine-tuning approaches using data-reuse and regularization.

\subsection{Reusing old data}
\label{sec:old_data}

We consider two main ways to reuse the old data ($D_1$) during fine-tuning: \textbf{replay} and \textbf{sampling}.
Experience replay \cite{lin1992self,de2019episodic} of size $N$ saves $N$ training examples into the replay buffer, allowing reuse during the next continual learning iteration. Usually, $N$ is small, but we experiment with various $N$ ranging from 0 to the size of the dataset.
To better parametrize experience replay for our setup, instead of $N \in \mathbb{N}$ we select $p \in [0, 1]$ -- where $p$ is the fraction of $D_1$ dataset that is used during fine-tuning.

Note that experience replay only uses a part of the dataset.
This may be beneficial in some setups, but we expect that using the whole dataset can reduce forgetting even further. Thus, our second method samples batches from both ``old'' and ``new'' subsets. We weigh examples from the ``old'' and ``new'' data differently and always weight ``new'' data more.
To simplify comparison with experience replay, we also parametrize this method with $p$, where $p$ is an expected proportion of old dataset sampled by the end of the epoch.

\subsection{Regularization-based methods}

Another approach to mitigate forgetting is to regularize the model in a way so that the model's predictions on the old data are changed minimally.

\paragraph{Move Norm Regularization}

We introduce a simple regularization method similar to weight decay but one that explicitly targets incremental training / continual learning.
Move norm is a regularizer that prevents the model with weights $\Theta$ from diverging from the previous model weights $\Theta^{prev}$ (Equation~\ref{eq:move}).
It is added to the loss function analogous to weight decay and is parametrized by $\lambda$.

\begin{equation}
    R(\Theta, \Theta^{prev}) = \lambda || \Theta - \Theta^{prev} ||_2
\label{eq:move}
\end{equation}


\paragraph{Elastic Weight Consolidation}

EWC was introduced in \citet{Kirkpatrick2017} and can be described as a parameter-weighted move norm:

\begin{equation}
    R(\Theta, \Theta^{prev}) = \lambda || F \circ (\Theta - \Theta^{prev})) ||_2
\label{eq:ewc}
\end{equation}

where $\circ$ is element-wise multiplication, and $F$ is a vector of positive parameter importance values. Theoretically, $F$ is the inverse of the diagonal of the Fisher matrix, but in practice, the squared gradients of the model trained on $D_1$ are used as an approximation.

The issues of applying EWC to modern neural networks are known in the literature \cite{lee2017toward} and come from the very flat loss surface at the end of the training. To mitigate this, we use the average values of the squared gradients over training steps, starting at the very first step. This method is similar to \cite{zenke2017continual} and we found it to significantly improve the exponential smoothing of the gradient norm proposed by \citet{lee2017toward}.

\subsection{Sampling EWC}

Even though the original EWC formulation specifically targeted the case with no old data available, we propose to combine EWC with sampling/replay. We expect to see additional improvements in terms of test-set performance improvement and reduced amount of forgetting.

\section{Experimental setup}
\label{sec:setup}

The model used in the study is a sequence-to-sequence Transformer model with a
pointer network and BERT encoder, as in \citet{rongali2020don}. We use the same hyperparameters\footnote{Details in the Appendix \ref{apx:hparams}}. For fine-tuning with EWC we use regularization strength of $100$ (regular experiments) and $1000$ (``high EWC'' experiments).

To mimic an incremental learning setup we split the training set into two segments:
``old data'' ($D_1$)
and
``new data'' ($D_2$)
.
The splitting procedure aims to mimic the real-world iterative setup when a trained
model already exists and we want to incorporate new data into this model.

\subsection{Data splits}

We experiment with multiple data splits, summarized in Table \ref{tab:splits}.
For each split, we chose a split class $C$ and a split
percentage $P_C$. We then randomly select $P_C$ percentage of the training examples
with class $C$ for the ``new data'' subset.
We selected mid- and low-frequency classes in our experiments.
In such cases, a 95\%+ split leaves a handful of examples in the $D_1$ subset and moves the rest to $D_2$.

To ensure that the ``new data'' does not contain classes absent from ``old data'', we add a small set of examples that contain all the classes to the ``old data''.

\begin{table}[]
    \centering
    \scalebox{0.9}{
    \begin{tabular}{l|lll}
        \toprule
        Split                  & Dataset & \# old &\# new \\\midrule
        Name Event 95          & TOP     & 60     & 1.1k
        \\
        Path 99                & TOP     & 15     & 1.5k
        \\
        Organizer Event 95     & TOP     & 15     & 301
        \\
        Get Loc. School 95 & TOP     & 9      & 171 
        \\
        Get Weather 95         & SNIPS   & 95     & 1.8k
        \\
        Served Dish 90         & SNIPS   & 25     & 230
        \\\bottomrule
    \end{tabular}}
    \caption{
        Data splits used in the experiments, showing the number of examples of the target class contained in the ``old'' and the ``new'' segments of the data.
        The first column shows the class name, followed by the proportion of training examples moved to the the $D_2$ subset. 
        }
        
    \label{tab:splits}
\end{table}


\subsection{Datasets}

We use two popular task-oriented semantic parsing datasets: Facebook TOP~\cite{gupta2018top} and SNIPS~\cite{coucke2018snips}.
SNIPS consists of flat queries containing a single intent and simple slots (no slot nesting), while the TOP format allows tree structures.  More specifically, each node of the parse tree is an intent name (e.g., IN:GET\_WEATHER), a slot name (SL:DATE), or a piece of input query text.
The tree structure is represented as a string of labeled brackets with text, and the model predicts this structure using the input query in a natural language (see Fig. \ref{fig:tree_paths}).

The TOP dataset consists of 45K hierarchical queries, 
about 35\% of which have tree depth $>$ 2.
SNIPS dataset has a simpler structure (only a single intent for each query)
and the train part consists of 15K examples.
To unify the datasets, we reformat SNIPS to fit the TOP structure.

\subsection{Metrics}
\label{sec:metrics}

We propose a new metric for semantic parsing, Tree-Path $F_1$ score. This metric that evaluates the accuracy of ``subcommands'' in the parsing tree -- paths from the root to a leaf. It is more fine-grained than EM and allows to compute class-specific scores.

We use exact match (EM) to evaluate model performance across all classes and Tree Path $F_1$ score (TP-$F_1$)
to evaluate performance on a specific class.
To estimate the uncertainty of our metrics, we divide our test set into 5 folds and compute the mean and standard deviation of model performance across the folds.

\paragraph{Tree Path Score}
\label{sec:tpf1}

To compute Tree Path $F_1$ Score (TP-$F_1$), the parse tree is converted into tree paths from the root to node that is either a valid slot or an intent without slots.
Tokens that do not correspond to any slot value are ignored.
This procedure is performed for both correct and predicted trees and then the $F_1$ score
is computed on the paths.

With
precision $
    P = \frac
        {\# \text{correctly predicted paths}}
        {\# \text{predicted paths}}
$
and recall
$
    R = \frac
        {\# \text{correctly predicted paths}}
        {\# \text{expected paths}}
$
.

\begin{figure}
    \centering
    \includegraphics[width=0.47\textwidth]{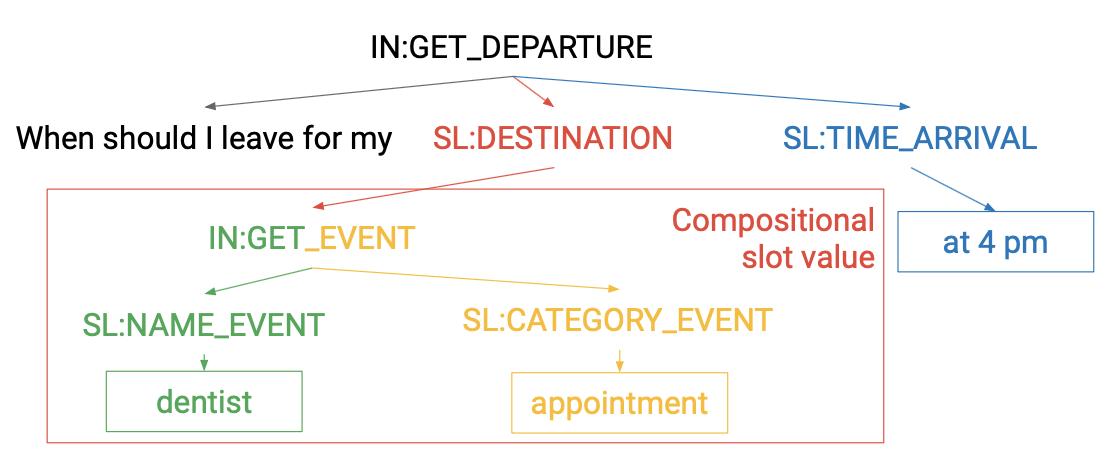}
    \caption{
    Semantic parsing example.
    Each tree path starts at the {\sc in:get\_departure} node and finishes with a slot value
    (the values are in the boxes).
    Tree paths are colored.
    }
    \label{fig:tree_paths}
\end{figure}

For example, for the query, ``When should I leave for my dentist appointment at 4 pm'', the parse tree looks as in Figure \ref{fig:tree_paths} and has four tree paths.
Every path starts at the root ({\sc in:get departure}) and goes down to the slot value.
For the {\sc sl:destination} slot, the value is compositional and equal to the
string
{\sc [in:get\_event [sl:name\_event: dentist] [sl:category\_event: appointment]]}
.

For example, if the predicted tree has a different value for the
slot {\sc sl:name\_event},
two paths in it would differ from the correct ones.
A path to the value of the {\sc sl:name\_event} slot and a path
to the value of the {\sc sl:destination} slot -- 
because {\sc sl:destination} value is compositional and contains
{\sc name\_event} as a part of it.
In this case, the number of correctly predicted paths would be two ({\sc sl:time\_arrival} and {\sc sl:category\_event} slots),
the number of predicted paths would be four, and the number of expected paths also four.
TP-$F_1$ in this case equals 1/2.

To compute a per-class score only the paths containing the target class are considered. In the example above, {\sc name\_event}'s TP-$F_1$ is zero, as there are no correctly predicted paths for it.

\newcommand{\ul}[1]{\underline{#1}}


\section{Results}
\label{sec:results}
\begin{table*}
\centering
\scalebox{0.9}{
    \begin{tabular}{l|cc|cccccccc}
    \toprule
    \multirow{2}{*}{Split} &
      \multicolumn{2}{c|}{from scratch} &
      \multicolumn{2}{c}{no finetunig} &
      \multicolumn{2}{c}{na\"ive} &
      \multicolumn{2}{c}{EWC} &
      \multicolumn{2}{c}{ours} \\
    & TP-F1 & EM & TP-F1 & EM & TP-F1 & EM & TP-F1 & EM & TP-F1 & EM\\
    \midrule
    Name Event 95
    & \bf{0.84} & \bf{0.82} & 0.73 & \ul{0.80} & 0.50 & 0.76 & 0.62 & 0.79 & \ul{0.81} & \bf{0.82} \\
    Path 99
    & \bf{0.64} & \bf{0.82} & 0.19 & 0.79 & 0.30 & 0.72 & 0.49 & 0.78 & \ul{0.63} & \ul{0.81} \\
    Organizer Event 95
    & \ul{0.60} & \bf{0.82} & 0.35 & \ul{0.81} & 0.23 & 0.78 & 0.52 & 0.80 & \bf{0.74} & \bf{0.82} \\
    Get Loc. School 95 \footnotemark{}
    & 0.65 & \bf{0.82} & 0.15 & 0.82 & \ul{0.66} & \ul{0.81} & 0.62 & 0.80 & \bf{0.79} & \bf{0.82} \\
    Get Weather 95
    & \bf{0.97} & \bf{0.95} & \ul{0.93} & 0.92 & 0.59 & 0.67 & 0.65 & 0.83 & \bf{0.97} & \ul{0.94} \\
    Served Dish 90 \footnotemark[\value{footnote}]
    & \ul{0.90} & \bf{0.95} & 0.81 & 0.92 & 0.59 & 0.90 & 0.84 & 0.91 & \bf{0.94} & \ul{0.94} \\
    \bottomrule
  \end{tabular}}
    \caption{
    Target class TP-F1 and exact match for the baselines and our methods. The baselines include training from scratch, fine-tuning the model on new data only without any regularization (na\"ive) and with EWC. For ``ours'' we select the best combination of method (Sec. \ref{sec:methods}) and old data amount. In all splits we are able to achieve comparable to a from scratch performance while na\"ive and EWC fail to do so in all splits except {\sc get\_loc\_school 95}.
    }
    \label{tab:performance}
\end{table*}

\begin{figure*}[h!]
    \centering
    \includegraphics[width=0.47\textwidth]{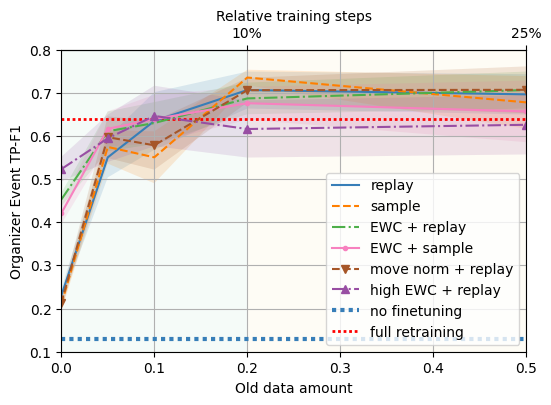}
    \includegraphics[width=0.47\textwidth]{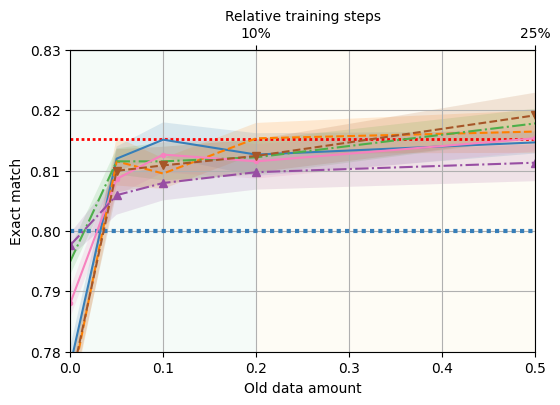}\\
    \caption{
    We plot class-specific TP-$F_1$ and exact match metrics against the fraction of original data sampled/replayed for the {\sc organizer\_event 95} split.
    Both sampling and replay from the old data help significantly.
    Similar performance to a full retraining can be achieved with sampling only around 10\% of the original data (using either of the methods). This comes at a fraction of computational cost of a training a new model.
    }
    \label{fig:organizer95_f1_em}
\end{figure*}

\subsection{Performance}


Achieving approximately the same as a trained from scratch model performance is an important requirement for the applicability of our method in practice.
To assess whether the proposed fine-tuning method performs on par with a model trained from scratch on a combined ``old''+``new'' dataset, we compare the following metrics: TP-F$_1$ for the target class and exact match.
The former evaluates the gain of using the additional data, the latter evaluates the model across all classes.

The Table \ref{tab:performance} shows that our methods that reuse old data
are able to achieve the performance 
of a fully retrained network.
At the same time, 
na\"ive fine-tuning sometimes performs \textit{worse than the current (not fine-tuned) model even on the target class}.
A popular continual learning approach, EWC, significantly improves upon na\"ive fine-tuning, but lags behind full retraining in performance.

The Figure \ref{fig:organizer95_f1_em} shows five different fine-tuning setups for the {\sc organizer\_event 95} split.
In each, we vary the amount of old data the method uses (parameter $p$, Section \ref{sec:old_data}).
We can see that na\"ive fine-tuning (the leftmost points for ``replay'' and ``sample'') fails to match the performance of full retraining even on the target class which is well represented in the fine-tuning data.
Same holds true for EWC without data sampling (the leftmost points for ``EWC+replay'', ``EWC+sample'', ``high EWC+replay''). Additional experimental results are provided in the Appendix \ref{apx:results}.

We also plot an auxiliary x-axis (top) that shows the number of training steps used for fine-tuning. To simplify the comparison with the model trained from scratch, we convert them to relative training steps. For the TOP dataset, 100\% is 28,000 steps that are needed to train a new model.

\begin{table*}[h!]
    \centering
    \scalebox{0.9}{
    \begin{tabular}{l|ccccc}
        \toprule
        Split
        & Replay & Sample & MoveNorm+Replay & EWC+Replay & EWC+Sample \\
        \hline
        Name Event 95
        & \bf{26\%}  & \bf{26\%} & \bf{26\%} & \bf{26\%}  & \bf{26\%} \\
        Path 99
        & 40\%  & \bf{29\%} &  40\%  & N/A        & N/A \\
        Organizer Event 95
        & 6\%   & 10\% &  10\%   & \bf{3\%}        & \bf{3\% }\\
        Get Loc. School 95 \footnotemark[\value{footnote}]
        & \bf{3\%}   & \bf{3\%} &  25\%   & 5\%        & 5\% \\
        Get Weather 95
        & \bf{10\%}   & \bf{10\%} & \bf{10\%}  & 28\% & \bf{10\%} \\
        Served Dish 95 \footnotemark[\value{footnote}]
        & \bf{16\%}  & \bf{16\%} & \bf{16\%} & \bf{16\%} & \bf{16\%} \\
        \bottomrule
    \end{tabular}}
    \caption{
    Relative number of training steps needed to reach the full retraining performance (within two standard deviations). 100\% training steps means full retraining. N/A means the performance has not been reached by either variation of the method. Different methods perform similarly.
    }
    \label{tab:finetuning_time}
\end{table*}

\subsection{Speedup}
\label{sec:speedup}

The main advantage that justifies using continual learning instead of training a new model from scratch, is the speedup.
In our experiments, we have observed a large and consistent decrease in training steps required to match the performance of a model trained from scratch.

The best improvement was a 30x speedup (in the number of training iterations) for the {\sc get location school 95} and {\sc organizer\_event 95} splits (Table \ref{tab:finetuning_time}).
It is worth noting that both of these splits model a more realistic scenario with a moderately-sized data patch (hundreds of examples) compared to the {\sc path 99} split (1.5 thousand examples).
Time-wise, in our setup we observed a twenty-fold reduction in training time and we believe this may be pushed even further using a separate device (GPU/TPU) for evaluation.

\subsection{Amount of forgetting}

Looking at Table \ref{tab:finetuning_time}, one can notice that there is no clear winner in this speed battle. All methods perform quite similar and it may be hard to decide which one to use.
However, another question worth asking is, ``how much does the model degrade after the fine-tuning?''.
While low exact match usually indicates catastrophic forgetting, high EM values may be misleading. A significant improvement over several classes can increase EM just enough to hide some degraded classes.

It is questionable to deploy a model that is better than the previous one overall, but performs significantly worse for some specific scenarios.
To quantify this, we evaluate the number of classes that degraded significantly. We define ``significant'' as more than $2\sigma$ where $\sigma$ is a standard deviation evaluated using bootstrapping (Section~\ref{sec:metrics}).

\begin{figure}
    \centering
    \includegraphics[width=0.47\textwidth]{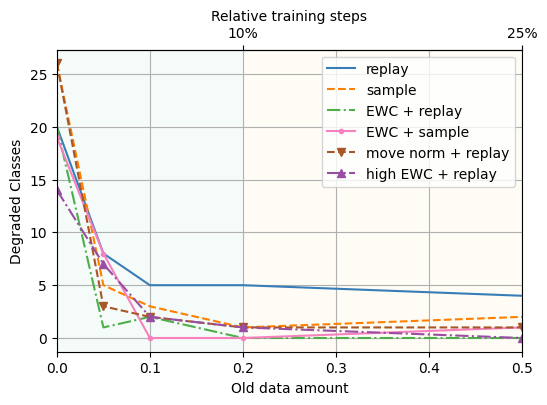}
    \caption{
    Number of classes with $2\sigma$-degraded tree path scores.
    }
    \label{fig:organizer95_forgetting}
\end{figure}

Figure \ref{fig:organizer95_forgetting} shows that the methods that use EWC and sampling exhibit less forgetting. For the case of {\sc organizer\_event 95} split EWC completely eliminates forgetting starting at 10\% of old data.

In summary, while different methods may yield similar results when looking at EM and class-specific TP-$F_1$ (Fig. \ref{fig:organizer95_f1_em}), they \textit{vary significantly in terms of the effects of catastrophic forgetting} (Fig. \ref{fig:organizer95_forgetting}). In all of our experiments we saw decreased amounts of forgetting when using EWC and old data sampling.
In 5/6 of our splits the method with the lowest forgetting was EWC+sampling.

\footnotetext{We observe high variance in metrics for the ``Served Dish~95'' and ``Get Loc. School 05'' splits. It could have been caused by a low number of test examples with the target class.}

\subsection{Negative Results: Layer Freezing}

One of the approaches we initially investigated was inspired by \citet{Howard2018}. Hypothetically, layer freezing during the fine-tuning stage could restrain the model from diverging too far and mitigate the forgetting. We experimented with freezing encoder/decoder/logits+pointer network and their combinations. In all experiments freezing performed worse than regular fine-tuning.
Freezing encoder and decoder and only updating the logits and pointer networks turned out to be a particularly bad method which suggests that improving low-resource classes that we targeted benefits mainly from constructing new features and not just reusing the learned ones.

\section{Related Work}
\label{sec:related_work}

Practical applications of Continual Learning \cite{mccloskey1989catastrophic,Parisi2019} to the NLP field has been rather limited \cite{li-etal-2019-compositional,lee2017toward}.
A usual task-incremental setup \cite{caccia2020online,de2019episodic} is far from the real-world applications.
However, a simpler data-incremental setup \cite{Cossu2020ContinualLW,mi2020continual} becomes extremely useful considering quickly growing size and computational requirements to train currently used NLP networks \cite{devlin2018bert,Radford2019LanguageMA,raffel2020exploring}.

The main issue of continual learning is catastrophic forgetting \cite{french1999catastrophic,mccloskey1989catastrophic}.
Data-based \cite{de2019episodic,Sun2020LAMOLLM} and regularization-based \cite{Kirkpatrick2017,zenke2017continual} methods has shown promising results in continual learning and applied to NLP tasks, including goal-oriented dialogue.
\citep{lee2017toward} applies continual learning methods for a goal-oriented dialogue task. Nonetheless, their setup is significantly different from ours as they do not thoroughly study the reduction of computational requirements using continual learning. Another difference is the model scale which is an important parameter in modern deep learning and NLP \cite{Kaplan2020ScalingLF}.
We argue that the effects of speedup in continual learning are the best motivation to study and apply these methods.

\section{Conclusion}

In this work, we consider a practical side of continual learning that has not received enough attention from the NLP researchers -- the ability to quickly update an existing model with new data.
Nowadays, training time becomes a more challenging issue every year.
We anticipate that in the near future of billion-parameter-sized models, incremental and continual learning settings can not only lead to a significant advantage in terms of resource efficiency but also become a necessity.

We demonstrate that in the data-incremental scenario, fine-tuning is a viable alternative to training a new model.
First, we show that simple data sampling techniques such as experience replay or supersampling allow to match the target-class and overall performance of a model trained from scratch.
Then, we demonstrate up to 30x speedup over training a new model.
Finally, we explicitly measure catastrophic forgetting and show that EWC and sampling from the whole dataset (opposed to experience replay) significantly reduce the forgetting effect.

We suggest using EWC with supersampling and monitoring specialized metrics like the number of degraded classes to track forgetting for the practical applications.
While different methods and different amounts of forgetting yielded the best results for different splits, EWC + 20\% sample generally performed well across all splits and registered low degradation.

To extend this work to a broader set of practical cases,
the future work will be concentrated on the class-incremental setup.
This will allow the additional data to have classes not represented
in the original dataset.

\bibliography{bibliography.bib}
\bibliographystyle{aaai}


\appendix
\section{Hyperparameters and training setup}
\label{apx:hparams}

Table \ref{tab:hyperparameters} contains hyperparameters used for pretraining.
We used the Noam schedule \cite{Vaswani2017} for the learning rate.
Note that it involves
$\frac{1}{\sqrt{d_{decoder}}} = \frac{1}{\sqrt{256}}$
learning rate scaling. 

For fine-tuning, the same parameters were used unless otherwise stated in the
experiment description.
We searched for the best EWC regularization from \texttt{[0.1, 10, 100, 100]} and for the best move norm in \texttt{[0.01, 0.05, 0.1]}.
If we say ``high EWC'' on our plots, it means EWC regularization factor to be $1000$.
Model, optimizer, and learning rate scheduler states were restored from the
checkpoint with the best EM.

\begin{table}[h]
    \centering
    \begin{tabular}{c|c}
        \hline
        encoder model & BERT-Base cased
        \\\hline
        decoder layers & 4
        \\\hline
        decoder n heads & 4
        \\\hline
        decoder model size & 256
        \\\hline
        decoder ffn hidden size & 1024
        \\\hline
        label smoothing & 0.0
        \\\hline
        batch size & 112
        \\\hline
        optimizer & ADAM
        \\\hline
        ADAM betas & $\beta_1=0.9$, $\beta_2=0.98$
        \\\hline
        ADAM $\epsilon$ & $10^{-9}$
        \\\hline
        decoder lr & 0.2
        \\\hline
        encoder lr & 0.02
        \\\hline
        warmup steps & 1500
        \\\hline
        freezed encoder steps & 500
        \\\hline
        dropout & 0.2
        \\\hline
        early stopping & 10
        \\\hline
        ranodm seed & 1
        \\\hline
    \end{tabular}
    \caption{Training and model hyperparameters}
    \label{tab:hyperparameters}
\end{table}


\section{Additional Experimental Results}
\label{apx:results}

\paragraph{Additional plots}

In this section we present additional experimental data, analogous to the one presented in the paper.

These plots exhibit a similar to their counterparts from Section \ref{sec:results} behavior.

\begin{figure}
    \centering
    \includegraphics[width=0.47\textwidth]{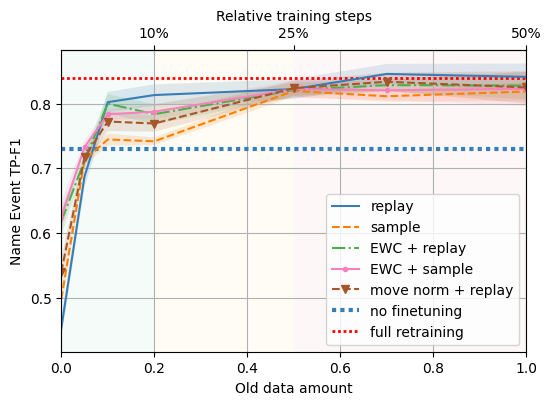}
    \includegraphics[width=0.47\textwidth]{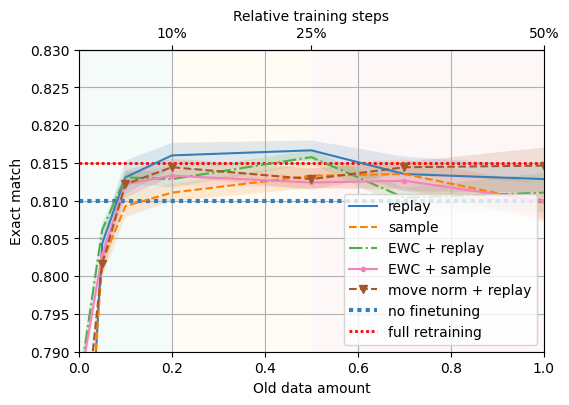}
    \includegraphics[width=0.47\textwidth]{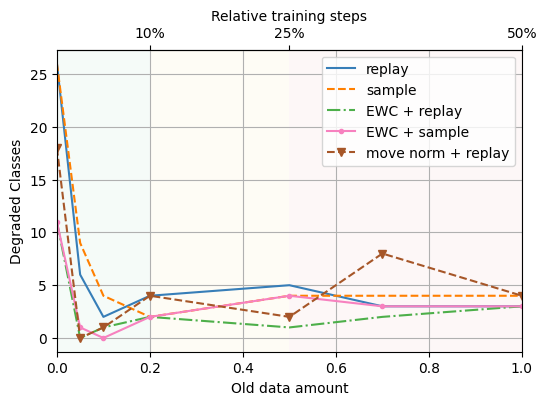}
    \caption{
    TOP NAME EVENT 95 results.
    }
    \label{fig:name_event95}
\end{figure}

\begin{figure}
    \centering
    \includegraphics[width=0.47\textwidth]{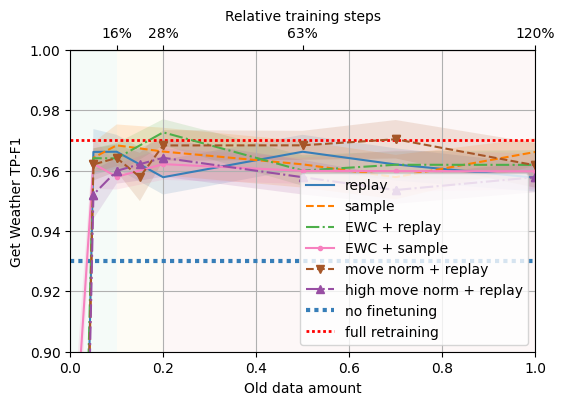}
    \includegraphics[width=0.47\textwidth]{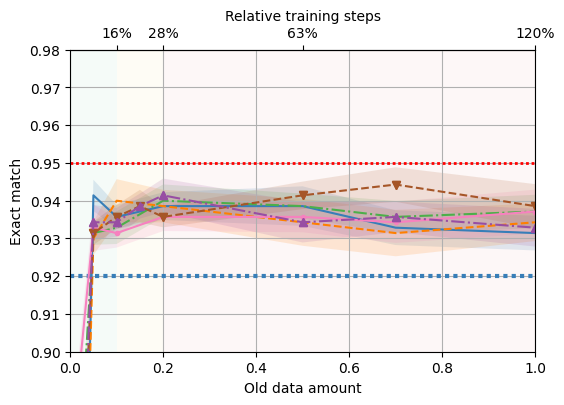}
    \includegraphics[width=0.47\textwidth]{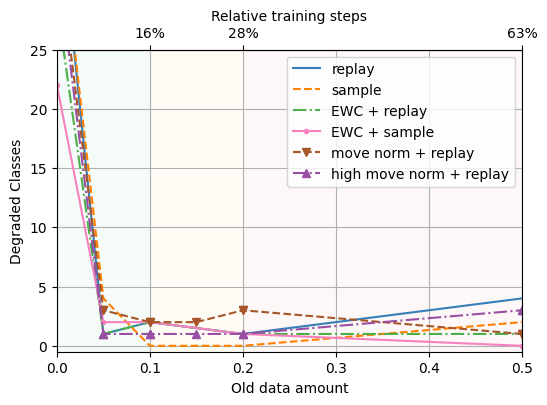}
    \caption{
    SNIPS GETWEATHER 95 results.
    }
    \label{fig:getweather90}
\end{figure}

\begin{figure}
    \centering
    \includegraphics[width=0.47\textwidth]{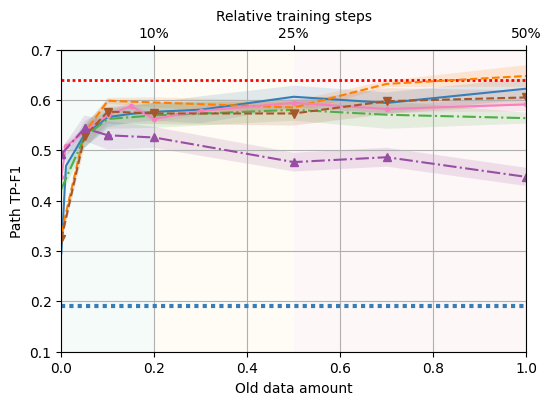}
    \includegraphics[width=0.47\textwidth]{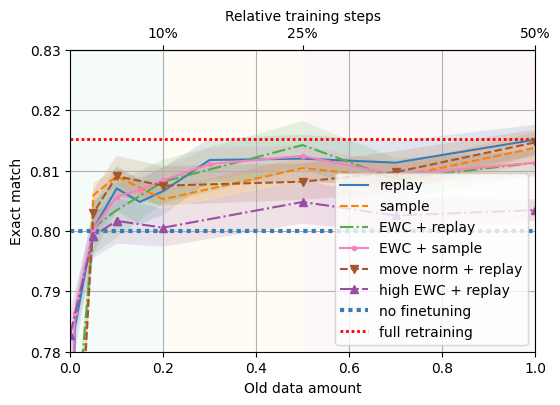}
    \includegraphics[width=0.47\textwidth]{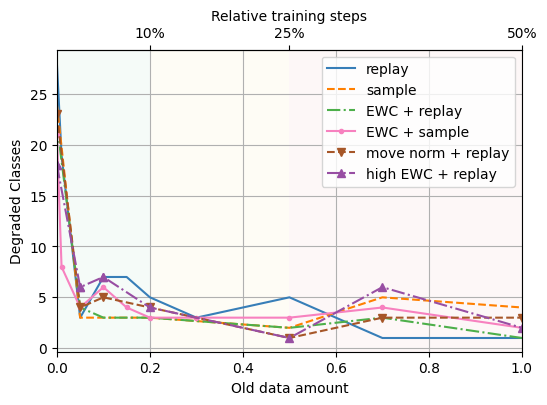}
    \caption{
    PATH 99 results.
    }
    \label{fig:path}
\end{figure}

\end{document}